\UseRawInputEncoding
\documentclass[runningheads]{llncs}

\usepackage{eccv}

\usepackage{eccvabbrv}

\usepackage{graphicx}
\usepackage{booktabs}
\usepackage{tabularx}
\usepackage[accsupp]{axessibility}  

\usepackage{lipsum}

\usepackage{hyperref}

\usepackage{orcidlink}
\usepackage{multirow}

\begin{document}

\title{An Intermediate Fusion ViT Enables Efficient Text-Image Alignment in Diffusion Models} 

\titlerunning{Abbreviated paper title}


\author{Zizhao Hu\inst{1}\and
Shaochong Jia\inst{1}\and
Mohammad Rostami\inst{1}}

\authorrunning{Z. Hu et al.}

\institute{University of Southern California, Los Angeles, CA, USA\\
\email{zizhaoh@usc.edu, jiashaoc@usc.edu, rostamim@usc.edu}}

\maketitle

\begin{abstract}
Diffusion models have been widely used for conditional data cross-modal generation tasks such as text-to-image and text-to-video. However, state-of-the-art models still fail to align the generated visual concepts with high-level semantics in a language such as object count, spatial relationship, etc. We approach this problem from a multimodal data fusion perspective and investigate how different fusion strategies can affect vision-language alignment. We discover that compared to the widely used early fusion of conditioning text in a pretrained image feature space, a specially designed intermediate fusion can: (i) boost text-to-image alignment with improved generation quality and (ii) improve training and inference efficiency by reducing low-rank text-to-image attention calculations. We perform experiments using a text-to-image generation task on the MS-COCO dataset. We compare our intermediate fusion mechanism with the classic early fusion mechanism on two common conditioning methods on a U-shaped ViT backbone. Our intermediate fusion model achieves higher CLIP Score and lower FID, with $20\%$ reduced FLOPs, and $50\%$ increased training speed compared against a strong U-ViT baseline with an early fusion.
\end{abstract}

\section{Introduction}
\label{sec:intro}
Diffusion models ~\cite{original-diffusion-paper} have emerged as a potent framework for generating high-definition images ~\cite{latent-diff, ddpm, beatgandiff, dalle}. Most existing methods leverage prealigned text embeddings such as CLIP (Dall-E \cite{dalle}, Dall-E 2 \cite{dalle2}, Dall-E 3 \cite{dalle3}, Stable Diffusion \cite{stable-diffusion}, Stable Diffusion XL \cite{podell2023sdxl})  
 and use concatenation or cross-attention to fuse the conditioning text embedding to the main image diffusion features. The core issue of this approach is the inherit gap between image and text representations in CLIP, plus the loss of specificity in certain contexts. Therefore, CLIP struggles in abstract tasks such as counting, in several types of fine-grained classification, and in out-of-distribution data \cite{clip}. All of these discussed features are essential to be learned in the diffusion process. Thus in the diffusion model, additional mechanisms are needed to align the text embedding to the image diffusion task. The exact location to fuse them should also reflect the natural information discrepancy between these two data types. However, existing methods lack such considerations. Some use simple concatenation or addition to fuse frozen text embeddings at the diffusion input, while others use only a single cross-attention layer to fuse the embeddings at all levels of the image diffusion model. These approaches directly inherit the limitations of CLIP embeddings, potentially damage the model's performance on the diffusion learning task, and introduce redundant text guidance with additional computing complexity. 

To enable better and more efficient text-image alignment, we bring the concepts from multimodal fusion to diffusion model conditioning and study how different fusion techniques will impact the model's performance on text-to-image generation tasks. We investigate a specific type of ViT-based segmentation backbones for the diffusion task. By examining the text-to-image and image-to-image attention maps at different layers at all time stamps, we discover that semantic information from captions provides more guidance at bottleneck layers whereas fusing captions at earlier or later layers provides minimum guidance. Based on this observation, we introduce an intermediate fusion mechanism for ViT-based diffusion backbones. This method removes the transformer blocks from both beginning and end layers and adds additional text-only transformer layers. Further analytical experiments indicate that this method improves the efficiency of text-image cross-attention mechanism. Our method enhances the efficiency of the remaining cross-attentions near the intermediate layers, leading to a better text-aligned, less interfered, and efficient image diffusion model. 

\begin{figure}[ht]
  \centering
\includegraphics[width=\linewidth]{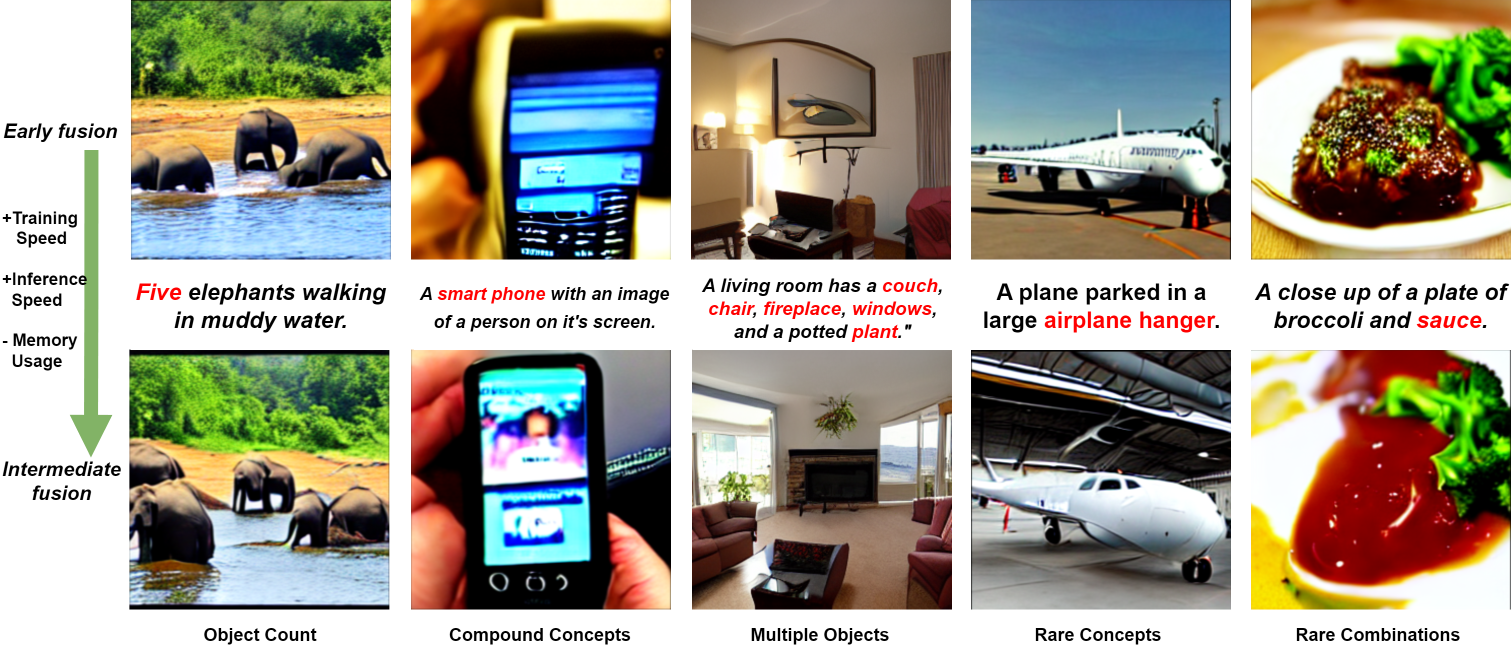}
  \caption{Text-to-image data generation using our intermediate fusion mechanism and the classic early fusion mechanism: images are generated based on a text input using the corresponding mechanism in each row. The input text contains a type of high-level semantics in each column. We observe that our mechanism generates better quality samples that align better with the text query while also being more efficient.}
  \label{fig: highlight}
\end{figure}

As a result, our proposed intermediate fusion enhanced backbone can generate better quality and more text-aligned images, especially for high-level semantics such as accurate object count, compound concepts, relationships between multiple objects, out-of-distribution generations including rare concepts and rare combinations of objects, etc.(see Figure \ref{fig: highlight} for a few generated samples). Our model is also more efficient compared to the commonly used early fusion mechanism in that it requires less computing, memory, training time, and inference time.

\section{Background and Related Work}

ViT-based flow-based model backbones have been explored recently, with significant success in DiT~\cite{peebles2023scalable} and U-ViT \cite{uvit}. They bring several large-scale applications in the text-guided generation domain (sora \cite{karaarslangenerate}, stable diffusion 3 \cite{stable-diffusion}) and multimodal generation domain (uni-diffuser \cite{unidiffuser}). These models leverage different mechanisms to fuse the text with the generation modalities, with simple concatenation and cross-attention being the two commonly used mechanisms. Most existing works introduce the text as the pretrained embedding features, which can be considered as early fusion in a diffusion model setting. Recent work suggests removing text embedding from the last few layers will reduce model FLOPs without affecting the model performance~\cite{peebles2023scalable}. However, this effect has not been studied systematically. As for text-image alignment, most existing works try to improve the alignment from a training perspective including finetuning with augmented data~\cite{paiss2023teaching, dalle3}, introducing additional alignment guidance~\cite{wu2023human}, etc. Quite differently, we approach addressing this problem from an architectural perspective to enable better cross-modal fusion. We briefly introduce the relevant background of our experiment setting.


\subsection{Diffusion Models}
Score-based diffusion models~\cite{song2020score, ddpm} are a class of generative models that learn to generate new samples from a target data distribution $p(x)$ by approximating the score function using a neural network. The score function provides a way to guide the generation of data without directly calculating complex probability normalization. It also mitigates the risk of mode collapse, encouraging a more diverse set of outputs that more accurately reflects the complexity of the target distribution. The score function is defined as the gradient of the log-probability density of the training data points:
\begin{equation}
S(x) = \nabla_x \log p(x).
\end{equation}

The training objective in score-based diffusion models is to minimize the discrepancy between the true score function and its neural network approximation. This is achieved by reducing the mean square error (MSE) between the true score function, $S(\cdot)$, and the approximation provided by the neural network, $s_\theta(\cdot)$:
\begin{equation}
L(\theta) = \mathbb{E}[(S(x_t) - s_\theta(x_t))^2].
\end{equation}

This modeling of the score function enables the model to generate the data distribution from a noisy distribution by iteratively adjusting its parameters in the direction of the gradient. This process involves taking small steps along the gradient direction to optimize the score function approximation.

\subsubsection{Training.}
Given a dataset $\mathcal{D}$ of independently and identically distributed (i.i.d.) real samples from the data distribution $p(x)$, the training process of the score-based diffusion model involves two steps:

\textbf{ (i) Noise addition.} Given a data sample $x$, we progressively inject Gaussian noise over $T$ steps until a noisy observation $x_T$ of the input data is generated. This process can be formalized as follows:

\begin{equation}
x_t = \sqrt{1 - \beta_t} x_{t-1} + \sqrt{\beta_t} \epsilon_t,
\end{equation}\\
where $\epsilon_t \sim \mathcal{N}(0, I)$, $t=1,\ldots,T$, $x_0 = x$, and $\{\beta_t\}_{t=0}^{T}$ is a noise schedule.

\textbf{(ii) Score function learning.} The   function $s_\theta(x_t, t)$ approximates the gradient of the log-density of the distribution $p(x_t | x)$. Let $S_t = s_\theta(x_t, t)$ be the approximated score function at time $t$,
and $G_t = \nabla_{x_t} \log p(x_t | x)$ be the gradient of the log probability with respect to $x_t$. Then, the loss function $\mathcal{L}(\theta)$ can be rewritten as:
\begin{equation}
\mathcal{L}(\theta) = \mathbb{E}_{p_{data}(x)}\left[ \frac{1}{2T} \sum_{t=0}^{T-1} \| S_t - G_t \|^2 \right].
\end{equation}

\subsubsection{Sampling.}
The sampling (inference) process of the score-based diffusion model is a noise addition process in a reverse direction:

\textbf{(i) Initial noisy observation.} We first initialize the noisy sample $x_T$ from a simple noise distribution, such as a standard Gaussian distribution.

\textbf{(ii) Denoising process.} We then apply the learned score function to denoise $x_T$ over $t \in [T,1]$ steps to gradually remove noise and generate a  sample:
\begin{equation}
x_{t-1} = \frac{1}{\sqrt{1 - \beta_t}} (x_t - \frac{\beta_t} {\sqrt{1-\bar{\beta_t}}}s_\theta(x_t, t)) + \sqrt{\beta_t}\epsilon,
\end{equation}

\subsection{Guided diffusion models}

Guided generation of data involves creating new data samples, $x$, based on given pair of data $(x_0, y_0)$. We can condition the score function on a label $y$ to learn conditional generative models for class-specific data generation. This objective can be done by leveraging the Bayes' rule to decompose the conditional score function~\cite{beatgandiff}:
\begin{equation}
\begin{split}
S(x_t|y) &= \nabla_{x_t} \log p(x_t|y) = \nabla_{x_t} \log p(y|x_t) + \nabla_{x_t} \log p(x_t)
\end{split}
\label{gfm}
\end{equation}

\textbf{Classifier guidance.} By providing a weight term $\lambda$ to adjust the balance between the unconditional score function and the classifier guidance during the sampling phase, we can recast Eq. \ref{gfm} as follows:
\begin{equation}
\begin{split}
S(x_t|y) = &\, \lambda \underbrace{\nabla_{x_t} \log p(y|x_t)}_{\text{classifier guidance}} + \underbrace{\nabla_{x_t} \log p(x_t)}_{\text{unconditional score function}}
\end{split}
\label{eq: cg}
\end{equation}
where the first term can be learned by a classifier.

\textbf{Classifier-free guidance (CFG).} We can also model the unconditional score function $\nabla_{x_t} \log p(x_t)$ and the joint score function $\nabla_{x_t} \log p(x_t, y)$ simultaneously to substitute the classifier guidance~\cite{ho2021classifier} for  obtaining a trade-off between the quality and   diversity of  samples. We replace $\lambda$ in Eq. \ref{eq: cg} with $1 +\omega$ to be consistent with the range $[0,+\infty]$ of the guidance scale $\omega$ proposed in the original paper:
\begin{equation}
\begin{split}
S(x_t|y) =  &\, (1+\omega) \nabla_{x_t} \log p(x_t,y) 
- \omega \nabla_{x_t} \log p(x_t) 
\end{split}
\label{eq: cfg}
\end{equation}
The conditional generation in our study is achieved through classifier-free guidance from caption $y$ to image $x$, while unconditional generation uses the same approach with an empty caption embedding.

\subsection{Latent Diffusion Models}
Latent Diffusion Models (LDMs)~\cite{stable-diffusion} operate directly in the latent embedding space of pretrained image features. By working in a lower-dimension latent space, LDMs enable efficient generation and training. The formulation is similar to vanilla diffusion models, except that a pretrained encoder $\phi(\cdot)$ such as CLIP is used to map images to a latent space $z = \phi{(x)}$ and the generated image is reconstructed from the denoised latent variable $\hat{z}$ using the pretrained decoder $\hat{x} = \theta{(\hat{z})}$. LDMs help enable efficient high-definition sampling with less computational burden. 

\subsection{Conditioning Methods}
Our major contribution is to enhance the current conditioning approaches by replacing early fusion with intermediate fusion. In our empirical exploration, we mainly focus on two common conditioning methods used by SOTA text-to-image diffusion models:

\textbf{ (i) Concatenation.} This approach concatenates the text embedding with the image features and leverages the self-attention mechanism in subsequent layers to fuse and guide image diffusion learning. See Figure \ref{fig:backbone} top row left.

\textbf{ (ii) Cross-attention.} This approach avoids additional text tokens in the intermediate layer by adding the text-to-image cross-attention to the image-to-image self-attention at each fusion level. See Figure \ref{fig:backbone} top row right.

We only use these two common approaches to show that our method is applicable and effective on different conditioning approaches. Although Other conditioning methods have demonstrated effectiveness in recent text-to-image diffusion models, testing all of them is beyond our resources and scope. We contend that our approach can be seamlessly integrated to other conditioning, provided that the dimensionality of internal representations remains uniform.

\section{Proposed Methodology}


Diffusion models are similar to hierarchical VAE decoders \cite{vahdat2020nvae} in the aspect of decoding intermediate latent space
with a hierarchical generation process. Such hierarchy leads to an observation that spatial and detailed information at lower levels do not align with the higher-level semantics in image captions that summarize the context. Our key intuition is that within the diffusion process if text conditioning is forced to occur later at a layer primarily concerned with higher-level semantics, more resources will be devoted to achieving a match with high-level semantic content and better alignment will be achieved. As a result, lower-level layers will suffer less from the interference caused by the inherit gap between CLIP text and image representations. We consider addressing the such interference using intermediate fusion which centers around two core ideas: (i) additional trainable text layers can pre-align text with the main diffusion model and (ii) fusing these learnable embeddings at intermediate layers to simulate the natural semantic information density of language and visual data. 

\subsection{Intermediate Fusion}

In our study, we consider two competitive multimodal fusion strategies from a text-to-image latent diffusion model perspective: (i) early fusion, where text conditioning is fused with image at the beginning, and (ii) intermediate fusion, where such fusion occurs at intermediate level. Most existing diffusion models only use the early fusion strategy. In this method, both the text and image embeddings are pretrained, frozen and concatenated at the input level, as shown in Figure \ref{fig:backbone} bottom row left. To address the shortcomings of this approach, we devise a diffusion backbone that has separate dedicated trainable layers for each data. Then the layer outputs for text and image will be concatenated at intermediate level, as shown in Figure \ref{fig:backbone} bottom row right.

\subsection{Diffusion Backbone Model}

\begin{figure}
    \centering
    \includegraphics[width = \linewidth]{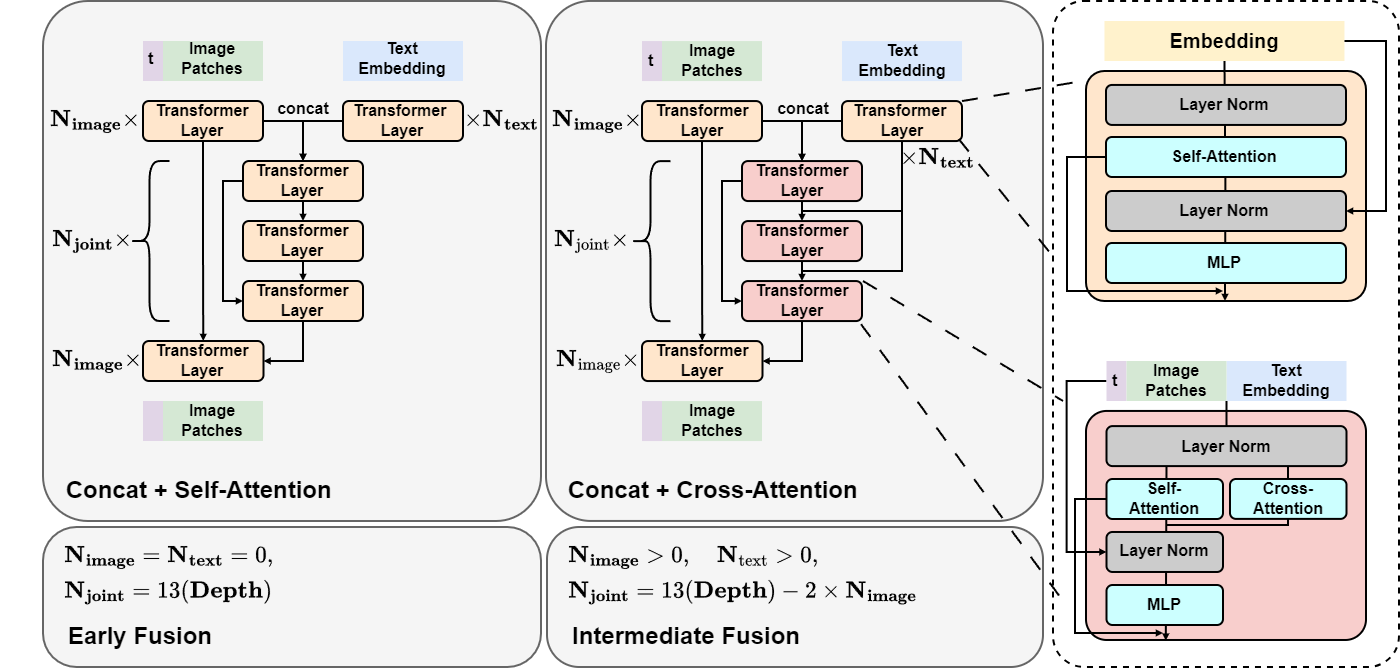}
    \caption{The two conditioning methods (top row) and two fusion methods (bottom row) in our experiments. The baseline is a 13-layer ViT model with the same skip-connection mechanisms introduced in U-ViT. For fusion comparison, we carefully designed $N_\text{joint}$ to ensure same depth and number of parameters for the image branch. In the experiments we use $N_\text{image} = 4$ and $N_\text{text} = 1$. For conditioning comparison, we switch the attention type and skip connections with other blocks intact.}
    \label{fig:backbone}
\end{figure}

 Evidence suggests that under a diffusion model setting, segmentation networks with long skip connections are essential to the efficient learning of discrete time ODE \cite{huang2024scalelong}. When long skip connections are used,  distant network blocks can be connected to aggregate long-distant information and alleviate vanishing gradient. 
  For this reason, we choose U-ViT-Small \cite{uvit} as our strong baseline. The baseline model uses self-attention with early fusion (Figure \ref{fig:backbone} entire left column). On top of this, we study three other settings, namely cross-attention with intermediate fusion (Figure \ref{fig:backbone} entire right column), self-attention with intermediate fusion, cross-attention with early fusion (Figure \ref{fig:backbone} diagonal combinations). These settings are constructed by only changing the backbone models without modifying the training and inference process.
  


\subsection{Latent Diffusion}

We use a stable diffusion KL-based autoencoder to encode an input image into the latent space and decode the denoised latent space representation back to the input image space. For text embeddings, we use the CLIP embedding with ViT-L-14. These models are frozen during diffusion model training.

\section{Experiments}
\subsection{Dataset and Training Settings}
 In our experiments, we use the MSCOCO \cite{coco} train and validation datasets to train and evaluate the performance of our model. For our training configuration, we train all models for 1 million steps and use a batch size of 256. We use the AdamW optimizer, with a learning rate of 0.0002, weight decay of 0.03, and beta parameters set to (0.9, 0.9). We incorporate a warm-up phase of 5000 steps to adjust the learning rate. The ViT model takes image features with a channel of 4, both spatial dimensions of 32, and an image patch size of 2. All attention mechanisms use an embedding dimension of 512, 8 attention heads. CLIP embedding has 77 tokens each with a dimension of 768, and is transformed to 512 dimension using a linear layer to align with the transformer input. For classifier-free guidance, we use a probability of 0.1 for unconditional training. For intermediate fusion, we use $N_\text{image} = 4$ and $N_\text{text} = 1$.
         
\subsection{Evaluation Metrics}

We deploy two evaluation methods, quantitative evaluation and human evaluation. Recent researches discover that FID cannot perfectly reflect the human preference for generation quality \cite{fid-pros-cons, fid-bad}, and CLIP Score cannot focus on specific alignment aspects such as matching object count \cite{clip}. Therefore, we also incorporate human evaluation for generation from specially designed prompts.

\textbf{(i) Quantitative evaluation.} We use FID, and CLIP Score as our quantitative metrics. To generate the score we select 30000 captions from the MSCOCO validation set and the corresponding generated images from our text-to-image models. For CLIP Score we use the CLIP version CLIP-ViT-L-14.

\textbf{(ii) Human evaluation - object count.} We choose a challenging generation aspect even for most of the foundation text-to-image models - matching object count, where we require the model to generate the same amount of objects as described in the prompts. We select four objects - bus(es), sheep, person(people), and apple(s). These four are selected since they represent 4 different plural forms and 4 categories(human, animal, fruit, human-made object). We use 5 words of count - a(an), two, three, four, five. Since larger numbers are rare in MS-COCO training captions, we restrict our study to small numbers. We generate 10 images for each object-count pair(20 pairs) and let evaluators count the number of target objects in the generated image. Then we use the average error and average match ratio to evaluate the performance: 

\begin{equation}
\text{Average\ Error} = \frac{\sum_{i=1}^{n} |\text{count}_{i,\text{human}} - \text{count}_{i,\text{prompt}}|}{n}
\end{equation}
\begin{equation}
\text{Average\ Match\ Ratio} = \frac{\sum_{i=1}^{n} \mathbb{I}(\text{count}_{i,\text{human}} = \text{count}_{i,\text{prompt}})}{n}
\end{equation}

\textbf{(iii)Human evaluation - preference score. } In addition, we ask 5 evaluators to provide a preference ranking for 1 to 4 on the overall quality of images generated by each model from random captions in the evaluation set. We use the same random seed and prompts for all models and provide the prompts and generated images to the evaluators. The human evaluators are allowed to skip any group of samples if the ranks are difficult to call. 100 captions are evaluated by 5 evaluators, with a maximum of 500 scores for each setting. We assign 4, 3, 2, and 1 scores to rank 1, 2, 3, and 4 respectively, and calculate the average score for each model setting.\\

\subsection{Results}
We selected U-ViT \cite{uvit} as our baseline model, since it has the best reproducible FID score among dedicated diffusion models trained on the MS-COCO dataset to our knowledge. We first compare the performance between these 4 experiment settings in Table \ref{tab:uvit_variants}.

\begin{table}[h]
\centering
\caption{Comparative results on text-to-image Generation and alignment metrics. The baseline U-ViT-Small correspond to the first settings. FID is evaluated on generated images from 30K captions from the evaluation set. CLIP Score is calculated on 30K generated samples and ground truth caption pairs using CLIP-ViT-L-14. For training speed, an iteration(iters) is a full forward-backward pass on an RTX-4090 GPU with a mini-batch size of 256. GFLOPs are calculated on a single forward pass of the model at one timestamp.}
\label{tab:uvit_variants}
\scriptsize
\newcolumntype{C}{>{\centering\arraybackslash}X}
\newcolumntype{L}{>{\raggedright\arraybackslash}X}
\newcolumntype{R}{>{\raggedleft\arraybackslash}X}
\begin{tabularx}{\textwidth}{@{}LLCCCCC@{}}
\toprule
Conditioning Method & Fusion Type & FID-30K $\downarrow$& CLIP Score $\uparrow$& Params & Training iters/s & GFLOPs \\
\midrule
\multirow{2}{*}{Concat} & Early(baseline) & 5.98 & 0.584 & 45M & 1.81  & 29.56 \\
 & Intermediate & \textbf{5.77} & \textbf{0.588} & 48M & \textbf{2.31} & \textbf{25.84} \\
\midrule
\multirow{2}{*}{Cross-attn} & Early & 6.48 & 0.575 & 45M & 2.54 & 23.82\\
 & Intermediate & \textbf{5.68} & \textbf{0.588} & 48M & \textbf{2.74} & \textbf{23.66}\\

\bottomrule
\end{tabularx}
\end{table}

\begin{figure}[ht!]
    \centering
    \includegraphics[width=\textwidth]{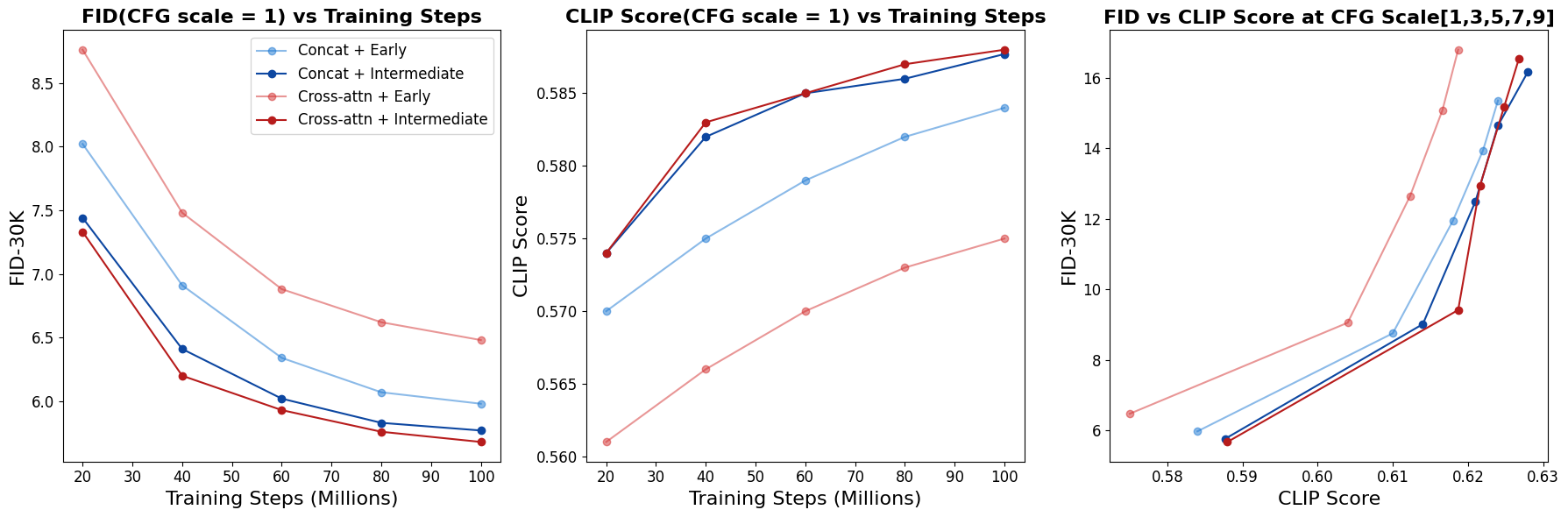}
    \caption{Evaluation during training and FID-30K vs CLIP Score at different CFG scales. Intermediate fusion settings show improved generation quality and text-image alignment compared to their early fusion counterparts. CLIP Score is measured on 30K pairs using CLIP-ViT-L-14.}
    \label{fig:three_metrics}
\end{figure}

In both conditioning methods, intermediate fusion shows lower FID and higher CLIP Scores than its early fusion counterpart. Compared to baseline setup, all other setups train faster and requires smaller GPU usage, even though some have more parameters. This is due to the reduced attention calculation. Next, we visualize the FID (Figure \ref{fig:three_metrics}, left), CLIP Score (Figure \ref{fig:three_metrics}, middle) during training, and FID vs. CLIP Score at different CFG scales (Figure \ref{fig:three_metrics}, right). 

From the results so far, We find that intermediate fusion settings improve all quantitative performance compared to their early fusion counterparts by a large margin. Among all four settings, an intermediate fusion with cross-attention has the best FID, CLIP score, lowest GFLOPs, and fastest training.

\begin{figure}[ht!]
  \centering
  \includegraphics[width=\linewidth]{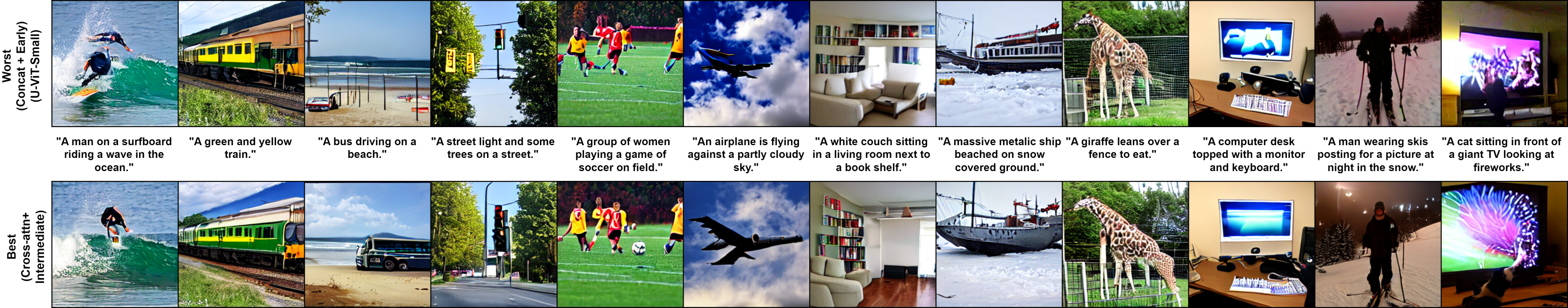}
  \caption{Comparison between the best intermediate fusion model(bottom) and the baseline model(top) across 12 different prompts. (Best viewed when zoomed-in.)}
  \label{fig:best-gen-vs-baseline}
\end{figure}

We select 12 random captions and generate images with a CFG scale of 3, which is an elbow point in the FID vs. CLIP Score curve. We show the baseline (top) compared with our best setting (bottom) in Figure \ref{fig:best-gen-vs-baseline}. 

We then show our best-performing setup against several foundation models and dedicated models (Table \ref{tab:model-performance}). Our model can reach the best text-image alignment performance and comparable image quality to all models with a relatively small model size.  

\begin{table}[htbp]
\centering
\caption{Performance of transformer-based text-to-image diffusion models.}
\label{tab:model-performance}
\scriptsize\newcolumntype{C}{>{\centering\arraybackslash}X}
\newcolumntype{L}{>{\raggedright\arraybackslash}X}
\newcolumntype{R}{>{\raggedleft\arraybackslash}X}
\begin{tabularx}{\textwidth}{@{}LLLCCC@{}}
\toprule
Model &Fusion Type & Fusion Path & FID-30K $\downarrow$& CLIP Score $\uparrow$& GFLOPs \\ \midrule
\multicolumn{6}{@{}l}{\textbf{Foundation Models Zero-shot on MS-COCO}}\\
\midrule
DALLE 3  & Early & Cross-attn & - & 0.320 & - \\
Imagen  & Early & Cross-attn & 7.27 & $\sim$0.29  & - \\
Stable Diffusion & Early & Cross-attn & 8.59 & ~0.325  & - \\
SDXL  & Early & Cross-attn &  & 0.305 & - \\
\midrule
\multicolumn{6}{@{}l}{\textbf{Dedicated Models Trained on MS-COCO}}\\
\midrule
U-ViT-Small  & Early & Concat & 5.98 & 0.584 & 29.56 \\
IF-U-ViT(ours)  & Intermediate & Cross-attn & \textbf{5.68} & \textbf{0.588} & \textbf{23.66} \\

\bottomrule
\end{tabularx}
\end{table}

\subsection{Human Evaluation}

\begin{figure}[ht!]
    \centering
    \includegraphics[width=\textwidth]{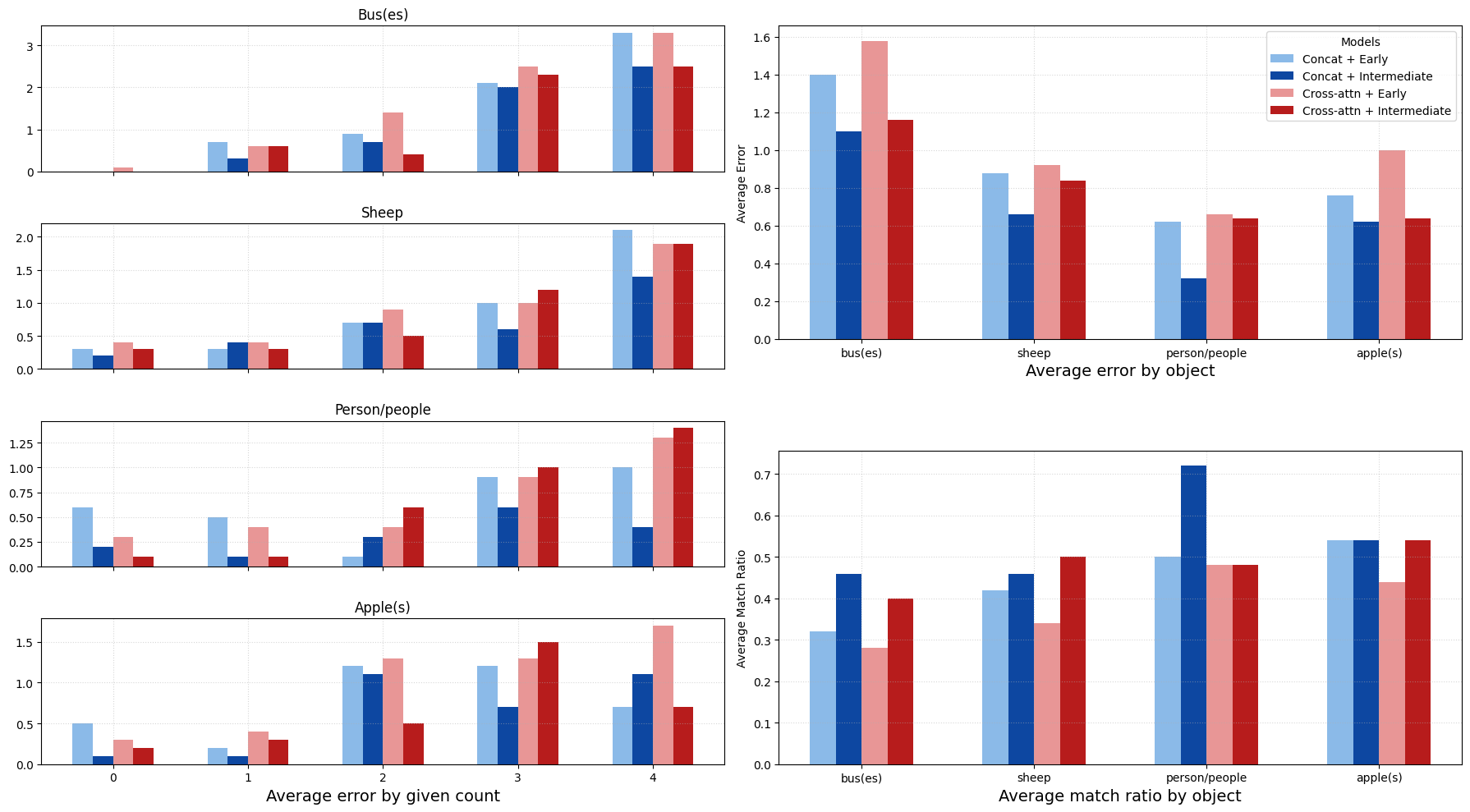}
    \caption{Human evaluation on object count. Lighter colors represent early fusion, while darker colors represent intermediate fusion. The left four figures are the average error given different ground truth counts, where x-axis is the ground truth. Each figure corresponds to an object. The right top figure is the average error across all counts for different objects. The bottom right figure is the average percentage of exact matches for each object. The plots indicated lower average count errors and higher matching counts of intermediate fusion.}
    \label{fig:count}
\end{figure}

\begin{figure}[ht!]
    \centering
    \includegraphics[width=\textwidth]{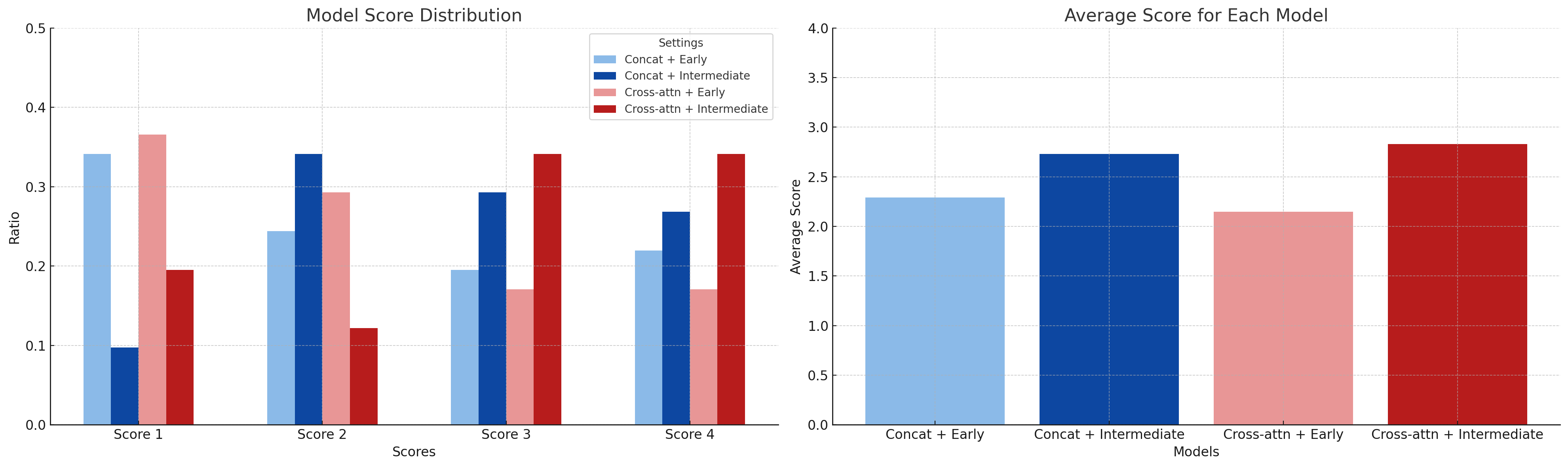}
    \caption{Human evaluation on general quality of generation.}
    \label{fig:quality}
\end{figure}

\textbf{Object count.} The results are shown in Figure \ref{fig:count}. In the left four figures, for 18 out of 20 object-count pairs, the intermediate concatenation generates objects with more human-aligned count compared to the early concatenation. For 14 out of 20 object-count pairs, the intermediate cross-attention generates objects with more or equal human-aligned count compared to the early cross-attention. In the top right figure, the average error of intermediate fusion is consistently lower than the early fusion counterparts. In the bottom right figure, the average match ratio of intermediate fusion is consistently higher or on par with the early fusion counterparts. All of the above results show that intermediate fusion improves the count alignment in the generation regardless of the conditioning method.
\\
\textbf{Preference score} The results are shown in Figure \ref{fig:quality}. 287 scores out of 500 expected scores are collected after removing invalid scores and those are too difficult to call by the human evaluators. The score is consistent with our FID and CLIP Score evaluation, with intermediate cross-attention achieving the highest score, and intermediate concatenation coming second. All intermediate fusion settings outperform their early fusion counterparts.

\subsection{Ablations}
\begin{figure}
    \centering
    \includegraphics[width=\linewidth]{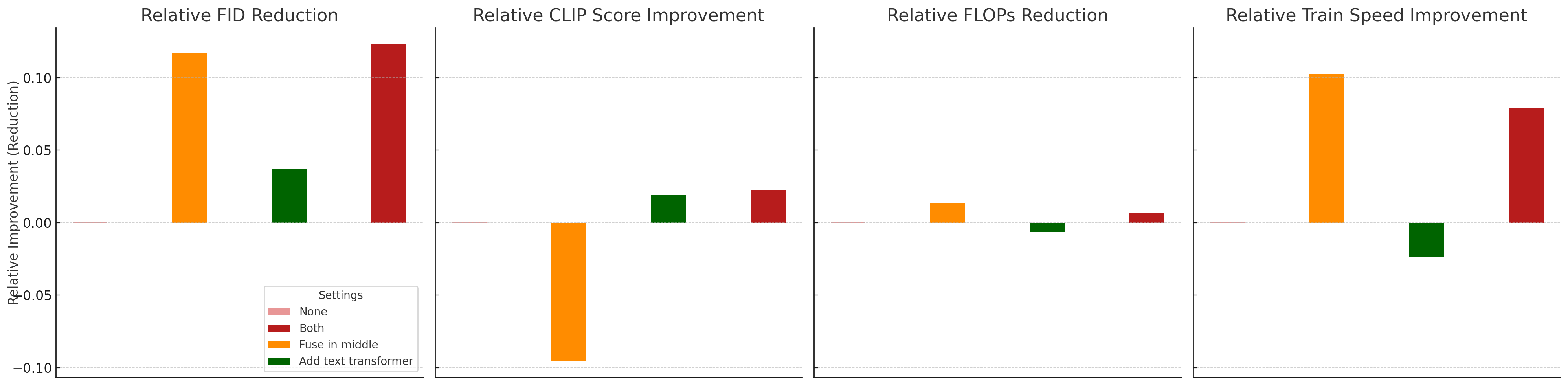}
    \caption{Ablation Study. We show the effect of adding trainable text transformer layers and fusing the embedding in the middle separately, together with the joint effect(Both) and baseline(None, early cross-attention). }
    \label{fig:ablation}
\end{figure}

Our intermediate fusion includes two atomic methods: 1. additional trainable transformer block for text embeddings during the image diffusion training 2. Only fuse these embeddings at the middle layers of a diffusion backbone. We study each of the method's contributions to the FID, and CLIP Score separately. In Figure \ref{fig:ablation}, we show that fusing the text embedding only in the middle of the diffusion backbone is the major source of FID improvement. This is expected since this setting has image-only skip connections that can maintain spatial consistency information at the upsampling layers. But this will negatively impact the CLIP Score. Adding text transformers learns more aligned text embeddings, which is the major source of improved CLIP Score, but this will impact the model efficiency in terms of increased FLOPs and reduced training speed. When the two methods are combined, we achieve improvements in all metrics. We see that these two atomic methods compensate for each other's weaknesses while maintaining their respective advantages.  

\subsection{Analysis}

\begin{figure}[ht]
  \centering
  \includegraphics[width=\linewidth]{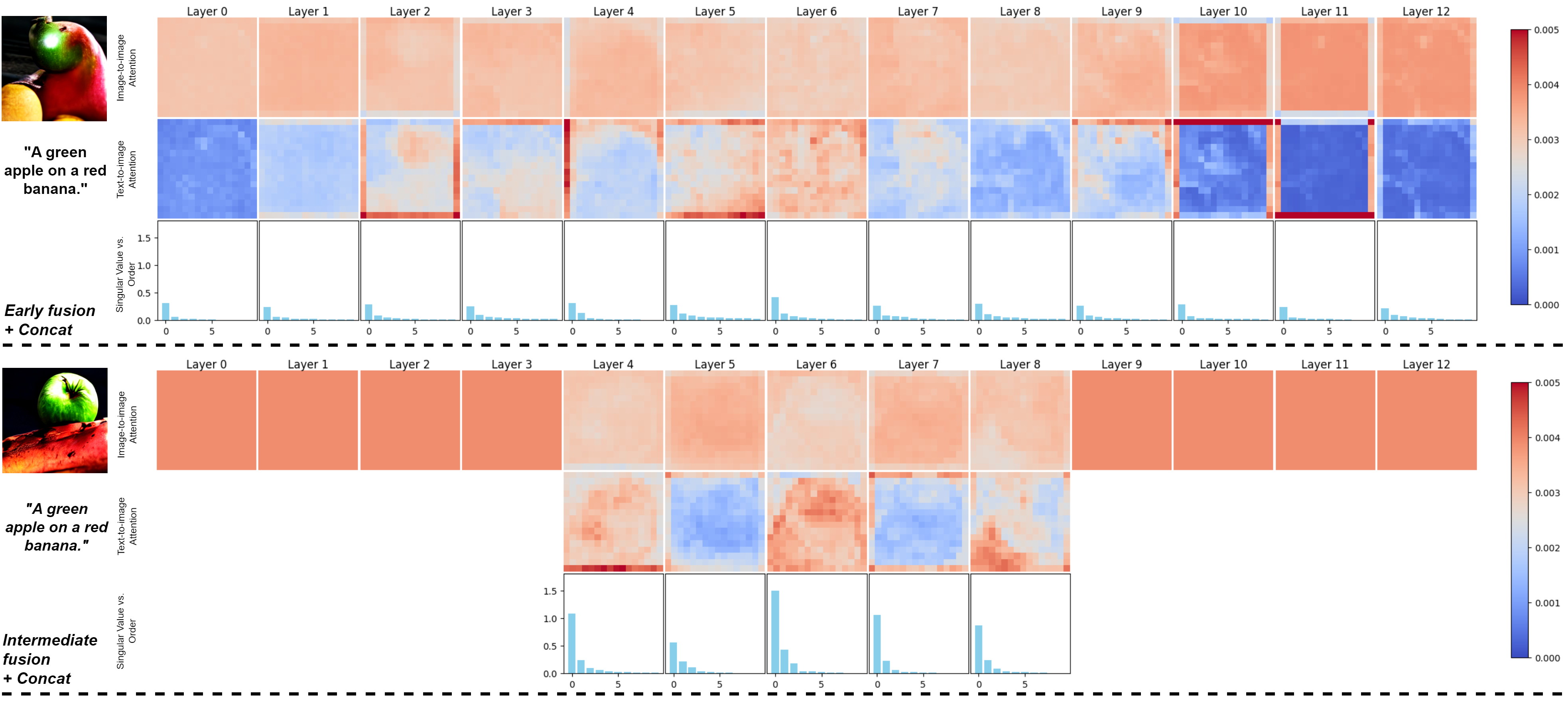}
  \caption{Attention maps and singular value analysis. For each setting, the first row is the image-to-image attention, the second row is the text-to-image attention, and the third row is the singular values of the first 10 orders from the text-to-image attention maps. An intermediate fusion removes the low-information text-to-image attention calculations at the early and late levels. This reduces the interference between image and text at these levels and improves the information capacity of the text-to-image attention at the middle layers. The combined effect results in better high-level semantic alignment and low-level spatial features' integrity. }
  \label{fig:4}
\end{figure}

\textbf{Layer-wise Attention Maps.}
To better visualize the text-image alignment across the model layers, we analyze the average attention maps of all timesteps during diffusion process.In Figure \ref{fig:4} we show the comparison of early and intermediate fusion. Attention maps of early fusion models provide a valuable observation.The text-to-image attention maps in both early and late layers indicate a more uniformly distributed pattern than intermediate layers, suggesting that text guidance is less focused and effective in early and late layers. Besides this observation, the early and late layers attend more to the border of the latent image due to the padded convolution encoding of images in the autoencoder model. To reduce the influence of such padded borders, we removed the border so that the later analysis can reflect the true image semantic guidance.\\


\textbf{Rank Analysis on Adjusted Attention Map }
To quantify the influence of text guidance on attention map image features, we conducted SVD on the attention matrix and analyzed their rank property. We see that early fusion models have relatively low-rank attention maps with smaller singular values at all layers, especially the layers away from the middle. The intermediate fusion on the other hand has high singular values. The analysis indicates that the early fusion introduces low-efficiency text-to-image attention in early and end layers of a ViT diffusion backbone, whereas most of the text information is fused around the bottleneck. This justifies our presumption of inconsistent alignment at lower levels of early fusion. Additionally, the comparison of quantified result proves that the elimination of the early and end fused layers never hurts the effectiveness of guidance. It instead boosted the guidance. Thus, by forcing the fusion to occur at a later stage, we can potentially improve model efficiency without damaging the semantic control of text. This observation aligns with the experiment results.

\section{Limitations}
\subsubsection{Other Conditioning Methods.}
The main focus of this paper is to investigate the influence of intermediate fusion provided by a new architecture backbone, in terms of text-image alignment, generation quality, and computational efficiency compared to early fusion models. Admittedly, early fusion covers a large number of conditioning strategies other than concatenation and cross-attention, and some methods such as the Adaptive Layer Norm used by DiT are not explored in this paper. We reasonably argue that the intermediate fusion can be transferred with ease to other unexplored conditioning strategies, since the approach resolved the issue of less efficient attention caused by joining image and text at input level.  \\

\textbf{Model Parameters and Hyperparameters.} Our model contains two hyperparameters $N_\text{image}$ and $N_\text{text}$(in latter discussion, a setting is denoted as [$N_\text{image}$, $N_\text{text}$]). In our experiments, we test one intermediate fusion setting([1,4]) in our main experiments and three settings([1,4],[0,4], and [1,0]) in ablation studies. However, how different architectural design will impact the results remains unexplored. This likely depends on the pretrained autoencoder, pretrained text embedding, embedding size, dataset, conditioning method, and other factors. Moreover, we logically contend that our approach can be applied to scaled-up foundation models, and will likely improve the performance with considerate design choices.

\section{Conclusion}
In this study, we presented an effective architecture for enhancing text-to-image diffusion models by leveraging an intermediate fusion mechanism for conditioning strategies. Our extensive experiments and analyses on the MS-COCO dataset demonstrate that this method significantly outperforms traditional early fusion techniques in aligning visual concepts with the high-level semantics of language. By integrating trainable text embeddings in the middle layers of a U-ViT backbone, our approach not only boosts the quality and text alignment of generated images but also enhances the efficiency of the training and inference. This improvement is quantitatively supported by improved CLIP Scores and reduced FID values, alongside qualitative assessments through human evaluations.

More generally, our findings suggest that the placement and integration of text embeddings within diffusion models play a critical role in the overall performance and efficiency of text-to-image generation tasks. By adopting an intermediate fusion strategy, we successfully alleviate the prevalent issue of misalignment between textual semantics and generated visual content. This provides a direction for large foundation models to a more efficient and text-aligned design.

%
%
\bibliographystyle{splncs04}
\bibliography{egbib}

\newpage
\appendix  
\section{Appendix}

This appendix provides additional details of the experiments and additional results discussed in the main paper. 

\subsection{Detailed Experimental Setup}
Provide detailed descriptions of the experimental setup that were not included in the main manuscript due to space constraints. Include any specific settings, parameters, or configurations used in your experiments.
\begin{table}[htbp]
\centering
\caption{Detailed Configuration of the models.}
\label{table:detailed_model_setup}
\resizebox{\textwidth}{!}{%
\begin{tabular}{l l l}
\hline
\textbf{Configuration Aspect} & \textbf{Parameter} & \textbf{Value} \\
\hline
\textbf{General Settings} & Random Seed & 1234 \\
                          & Tensor Dimensions & (4, 32, 32) \\
\hline
\textbf{Model Architecture} & Autoencoder Scaling Factor & 0.23010 \\
\hline
\textbf{Training} & Number of Steps & 1,000,000 \\
                  & Batch Size & 256 \\
\hline
\textbf{Optimization} & Optimizer Type & AdamW \\
                      & Learning Rate & 0.0002 \\
                      & Weight Decay & 0.03 \\
                      & Momentum (Betas) & (0.9, 0.9) \\
\hline
\textbf{Learning Schedule} & Scheduler Type & Linear \\
                           & Warm-up Steps & 5,000 \\
\hline
\textbf{Network} & Network Type & ViT with skip-connections\\
                              & Image Feature Size & 32 \\
                              & Input Channels & 4 \\
                              & Patch Size & 2 \\
                              & Embedding Dimension & 512 \\
                              & Network Depth & 12 \\
                              & Number of Attention Heads & 8 \\
                              & MLP Ratio & 4 \\
\hline
\textbf{Data Configuration} & Dataset & MSCOCO-256x256\\
                            & CFG Unconditional Probability & 0.1 \\
\hline
\textbf{Sampling Strategy} & Steps per Sample & 50 \\
                           & Total Samples & 30,000 \\
                           & Mini-batch Size & 50 \\
                           & Sampling CFG Scale & 3.0 \\
                           & Eval CFG Scale & 1.0 \\
\hline
\end{tabular}%
}
\end{table}

\subsection{Additional Results}
We present more text-to-image generations from 4 different models here. We use the same prompt randomly selected from the MSCOCO evaluation set with a classifier-free guidance level 3.

\begin{figure}
    \centering
    \includegraphics[width =0.9 \linewidth]{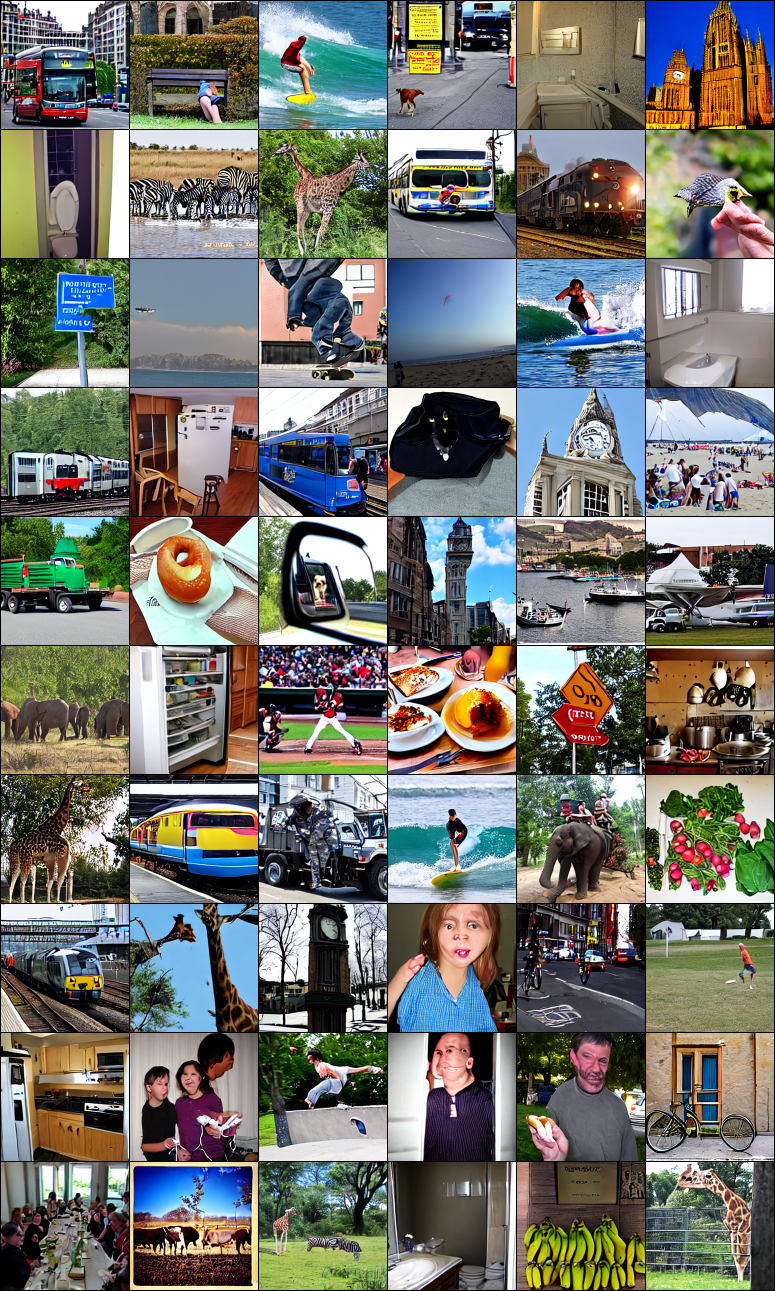}
    \caption{Early fusion + Concatenation(U-ViT-Small)}
    \label{Early fusion + Concatenation(U-ViT-Small)}
\end{figure}

\begin{figure}
    \centering
    \includegraphics[width =0.9 \linewidth]{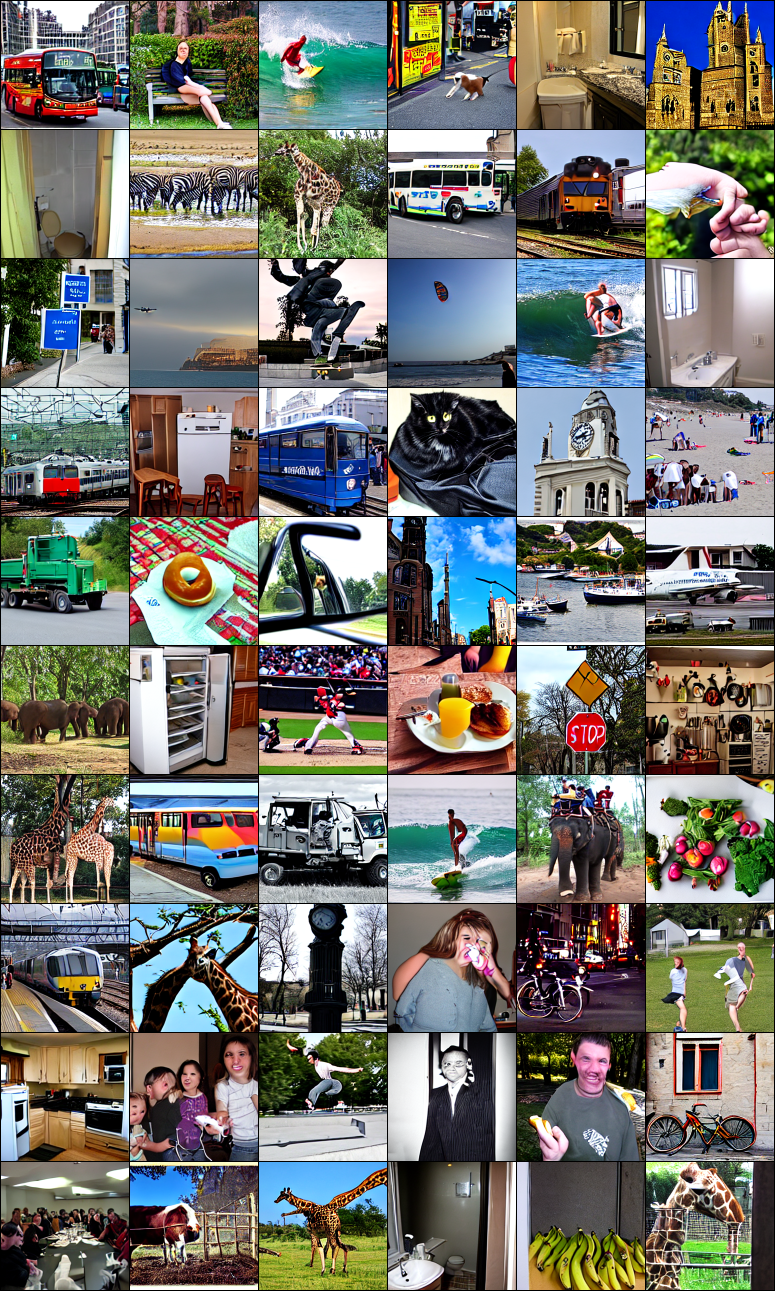}
    \caption{Intermediate fusion + concatenation}
    \label{intermediate fusion + concatenation}
\end{figure}

\begin{figure}
    \centering
    \includegraphics[width =0.9 \linewidth]{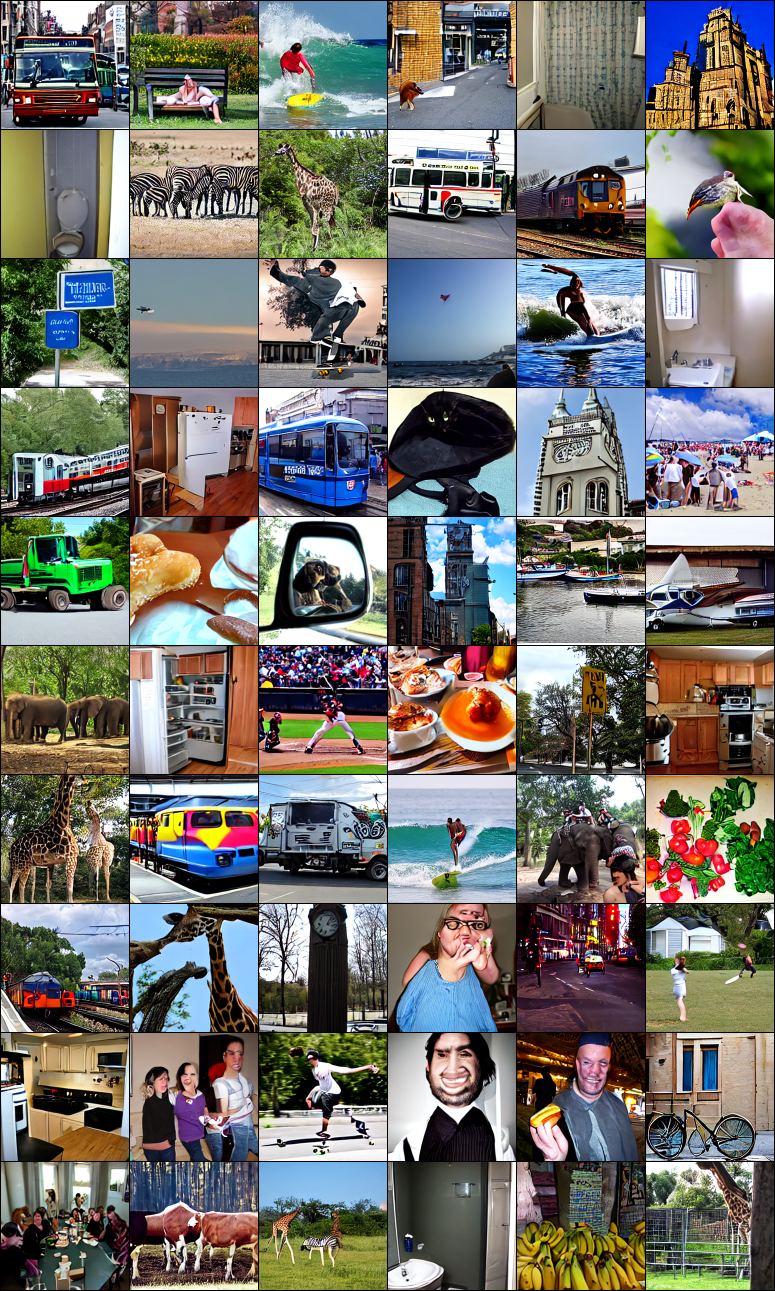}
    \caption{Early fusion + cross-attention}
    \label{Early fusion + cross-attention}
\end{figure}

\begin{figure}
    \centering
    \includegraphics[width =0.9 \linewidth]{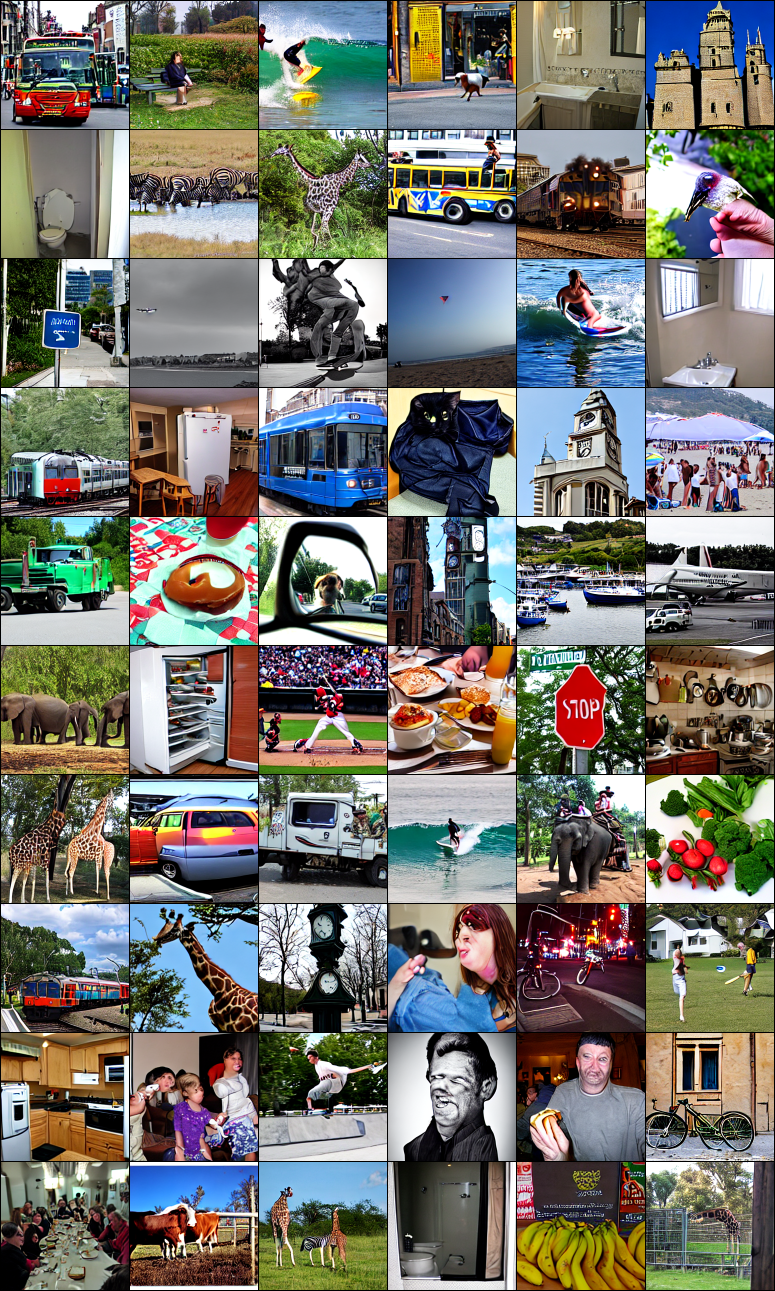}
    \caption{Intermediate fusion + cross-attention}
    \label{intermediate fusion + cross-attention}
\end{figure}

\subsection{Human Evaluation Setup}
We sent the following forms to 5 evaluators, together with 30 groups of generated samples for object count, and 100 groups of generated samples for preference ranking. The evaluators are not aware of the purpose of this project and the model identification throughout the evaluation. The model order is not randomized, so it's possible there are biases passed to later groups based on seen groups.
\begin{table}[htbp]
\centering
\caption{Human evaluation for object count form.}
\label{table:object_count}
\begin{tabular}{l@{\hspace{1em}}p{7cm}@{\hspace{1em}}r}
\hline
\textbf{ID} & \textbf{Question} & \textbf{Number of Objects Identified} \\
\hline
1 & How many \{Object\} can you see in the image? & 3 \\
\hline
\end{tabular}
\end{table}

\begin{table}[htbp]
\centering
\caption{Human evaluation for preference ranking form. The question is ``Rank the quality of the generation given \{Prompt\}.''}
\label{table:preference_ranking}
\begin{tabular}{l@{\hspace{1em}}r@{\hspace{1em}}r@{\hspace{1em}}r@{\hspace{1em}}r@{\hspace{1em}}c}
\hline
\textbf{ID} & \textbf{Image 1} & \textbf{Image 2} & \textbf{Image 3} & \textbf{Image 4} & \textbf{Skipped (Y/N)} \\
\hline
1 & 1 & 2 & 3 & 4 & N \\
\hline
\end{tabular}
\end{table}

\subsection{Code Availability}
The code will be released on GitHub.

\end{document}